\documentclass[runningheads]{llncs}
\usepackage{graphicx}
\usepackage{tikz}
\usepackage{comment}
\usepackage{amsmath,amssymb} % define this before the line numbering.
\usepackage{color}

\usepackage[accsupp]{axessibility}  % Improves PDF readability for those with disabilities.

\DeclareGraphicsExtensions{.pdf,.png,.jpg}
\usepackage[capitalize]{cleveref}
\crefname{section}{Sec.}{Secs.}
\Crefname{section}{Section}{Sections}
\Crefname{table}{Table}{Tables}
\crefname{table}{Tab.}{Tabs.}

\usepackage{booktabs}
\usepackage{adjustbox}

\begin{document}
% \linenumbers
\pagestyle{headings}
\mainmatter
\def\ECCVSubNumber{7396}  % Insert your submission number here

\title{HVC-Net: Unifying Homography, Visibility, and Confidence Learning for Planar Object Tracking} % Replace with your title

\begin{comment}
\titlerunning{ECCV-22 submission ID \ECCVSubNumber}
\authorrunning{ECCV-22 submission ID \ECCVSubNumber}
\author{Anonymous ECCV submission}
\institute{Paper ID \ECCVSubNumber}
\end{comment}
%******************

%\begin{comment}
\titlerunning{HVC-Net}
\author{Haoxian Zhang\inst{*}\inst{1}\orcidID{0000-0001-7078-868X},
Yonggen Ling\inst{*}$^\dagger$\inst{2}\orcidID{0000-0001-8294-6286}}
\authorrunning{H. Zhang, Y. Ling}
\institute{Tencent AI Lab, China \\
\email{leohxzhang@tencent.com}
\and
Tencent Robotics X, China \\
\email{rolandling@tencent.com}\\
\footnotetext{$^*$ Equal Contribution listed alphabetically. $^\dagger$ Corresponding author.}}
%\end{comment}
%******************
\maketitle

\begin{abstract}
Robust and accurate planar tracking over a whole video sequence is vitally important for many vision applications.
The key to planar object tracking is to find object correspondences, modeled by homography, between the reference image and the tracked image. 	
Existing methods tend to obtain wrong correspondences with changing appearance variations, camera-object relative motions and occlusions.
To alleviate this problem, we present a unified convolutional neural network (CNN) model that jointly considers homography, visibility, and confidence.
First, we introduce correlation blocks that explicitly account for the local appearance changes and camera-object relative motions as the base of our model.
Second, we jointly learn the homography and visibility that links camera-object relative motions with occlusions.
Third, we propose a confidence module that actively monitors the estimation quality from the pixel correlation distributions obtained in correlation blocks. 	
All these modules are plugged into a Lucas-Kanade (LK) tracking pipeline to obtain both accurate and robust planar object tracking.
Our approach outperforms the state-of-the-art methods on public POT and TMT datasets.
Its superior performance is also verified on a real-world application, synthesizing high-quality in-video advertisements.
%\dots
\keywords{Planar Object Tracking, Homography, Visibility, Confidence}
\end{abstract}

\section{Introduction}
\label{sec:intro}

Planar object tracking is a classic computer vision task with a wide range of applications.
Given the initial corners of a planar object in the reference frame, the primary goal of planar tracking is to estimate the movements of these corners, modeled by a geometric transformation called a homography, in consecutive frames.
Though lots of advances have been made in past decades, obtaining accurate and robust results remains challenging.
These difficulties are mainly caused by three factors: appearance variation, camera-object relative motion and occlusion.
\begin{figure}[!t]
	\begin{center}
		\includegraphics[width=1.0\textwidth]{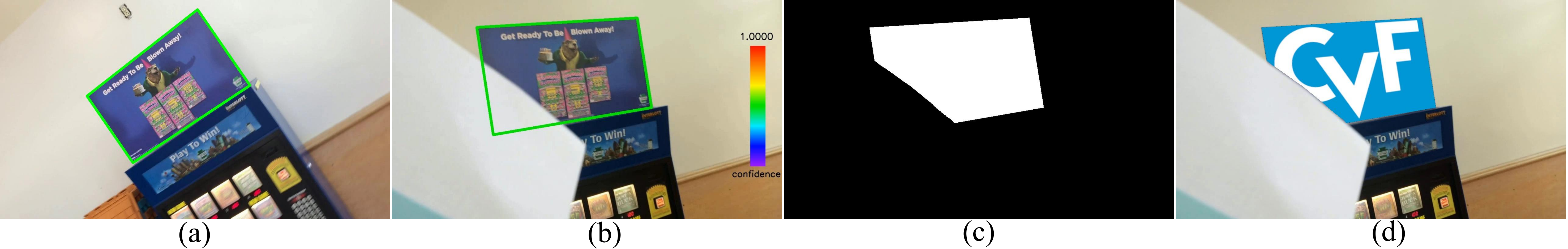}
	\end{center}
	
	\caption[aa]{One of our synthetic frames in the lottery sequence of \cite{planar_tracking_benchmark}.
		(a) A planar object to be tracked in the reference image, denoted by a green quadrilateral.
		(b) The estimated homography with very high tracking confidence in one video frame.
		(c) Corresponding visibility mask of the tracked object.
		(d) The synthetic frame after placing the CVF logo using results from (b) and (c).
		More results are shown in Fig.~\ref{fig:results} and the supplemental. %supplementary material.
	}
	\label{fig:spot}

\end{figure}
The appearance variation is a camera-related issue.
It is usually known as image blur, sensor noise, non-linear response of brightness.
The camera-object relative motion leads to geometry transformations of an object on the image.
Typical effects on the image plane are scale changes, rotations, translations, and perspective distortions.
Occlusion is referred as the fact that the tracked object is occluded by another object.
The situation becomes worse if the `another object' looks very similar to the tracked object.
These factors pose strong challenges for traditional keypoint-based methods that estimate the homography using hand-crafted features \cite{orb,sift,surf}, since the extracted features are prone to be different under the influence of these factors.
Learned features like D2-Net \cite{d2net}, LF-Net \cite{lfnet}, and R2D2 \cite{r2d2} are proposed to decrease this influence.
Direct methods \cite{baker2004lucas,esm04}, usually with the LK pipeline \cite{baker2004lucas}, estimate the homography iteratively.
\cite{baker2004lucas,esm04} assume the intensity consistency and compute the homography increment for each iteration.
\cite{clkn,nguyen2017unsupervised,lin2017inverse,zeng18rethinking,zhang2020content} extend direct methods with the learned `feature consistency' assumption for increasing the robustness.
We argue that efforts are still needed on better feature representation.
Moreover, these methods have not discussed occlusions that are widely existed in real-world video sequences.
The last to mention is the CNN-based method \cite{aniel16} that directly regresses the homography in one step with CNN.% the superpower of
It is not robust to these three factors, neither.

\begin{figure}[!t]
	\centering	
	\includegraphics[width=0.61\textwidth]{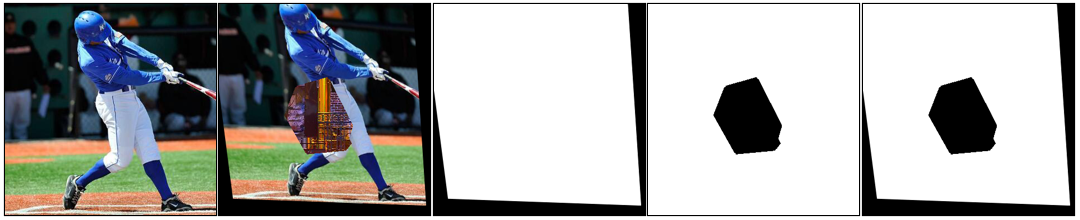}
	\caption{(from left to right) Reference image, current image, motion-induced visibility, disocclusion-induced visibility, combined visibility used in our model.
	}
	\label{fig:visibility}

\end{figure}

In this work, we propose a novel CNN model for handling mentioned difficulties.
The base of our model is correlation blocks (Sect.~\ref{subsec:decoupled_structure}).
It firstly extracts features in the intensity domain for handling appearance variations.
Cost volumes, representing distributions of pixel correlations, are then constructed in the pixel displacement domain to account for the camera-object relative motion.
We find that estimating the homography with these two cascaded steps is much better than methods with one step \cite{aniel16,nguyen2017unsupervised,clkn,lin2017inverse}.
Moreover, in contrast to methods that learn homography alone \cite{aniel16,nguyen2017unsupervised,clkn,lin2017inverse},  we learn it jointly with another task called visibility, which is defined as a binary mask that indicates which part of the reference image is visible on the tracked image (Fig.~\ref{fig:visibility}).
A reference image pixel is regarded as visible if and only if it satisfies the homography constraint of the tracked planar object (geometry-induced) and it is not occluded by other objects on the tracked image (disocclusion-induced).
Joint learning homography and visibility not only improves the correlation block representations, but also links camera-object relative motions with occlusions (Sect.~\ref{subsec:joint_learning}).
Lastly, as estimations with the LK pipeline are sensitive to initializations, we further improve the estimation robustness by monitoring the tracking quality and rebooting estimations.
This is done by introducing a confidence module that evaluates the planar tracking quality from pixel correlation distributions obtained in correlation blocks (Sec.~\ref{subsec:confidence}).
By equipping all these presented modules with a LK pipeline, our model obtains both accurate and robust homography estimations.
We achieve significantly higher homography precision than state-of-the-art homography estimation methods (Sect.~\ref{sec:exp}).
Besides, as a by-product, our model provides visibility masks that other works have not mentioned.
With these masks, we are able to easily place planar advertisements in videos (Fig.~\ref{fig:spot}).

\section{Related Work}

\subsection{Homography Estimation}
Existing planar tracking methods for estimating the underlying homography can be roughly classified into three categories: keypoint-based methods \cite{orb,sift,surf,FERN,BMIC,NNBMIC}, direct methods \cite{baker2004lucas,esm04,GO-ESM,SCV,L1APG,L1_track}, and CNN-based methods \cite{aniel16,nguyen2017unsupervised,clkn,lin2017inverse}.
Keypoint-based methods firstly detect and describe keypoints (using ORB \cite{orb}, SIFT \cite{sift}, SURF \cite{surf} and etc.) both in the reference planar region and subsequent consecutive frames.
These keypoints are then matched by minimizing the distances in the descriptor space.
Homography, the planar surface in the projection space is related, is then calculated with the obtained matches.
To remove potential outlier matches, RANSAC \cite{ransac} is usually performed.
Different from keypoint-based methods, direct methods \cite{baker2004lucas,esm04} assume that the planar template does not move fast in consecutive images.
The homography is directly optimized by minimizing the photometric error between the planar template and its projection in the incoming video frames.
Recently, CNN-based methods have been proposed.
Homography is regressed from input images in one forward step \cite{aniel16,nguyen2017unsupervised,zeng18rethinking,zhang2020content}.
\cite{clkn,lin2017inverse,Le_CVPR_2020,zhao2021deep} adopt the Lucas-Kanade framework \cite{baker2004lucas} and compute homography with multiple iterations.

\subsection{Object Segmentation}
The visibility of planar object tracking is less discussed in the past.
The closest work is segmentation.
There are three main approaches for object segmentation according to the level of supervision required.
Supervised methods require iterative human interactions for adding segmentation prior as well as refining segmentation outputs \cite{xuebai09,fan15tog}.
They obtain high-quality segmentations at the cost of extensive expert efforts.
To relax this mass manual supervision, semi-supervised methods propagate sparse human labeling in the reference frame to the remaining frames, and then formulate the segmentation problem as an optimization problem with energy defined over graphs \cite{vb17cvpr,sarv14cvpr,svkg12}.
The last to mention is the unsupervised methods that do not require any manual annotation or utilize prior information on the segmented objects.
Early unsupervised methods focus on over-segmentation \cite{mg10cvpr} or motion segmentation \cite{thomasbrox10eccv}.
They are extended to foreground-background separation in recent years \cite{wang15cvpr,taylor15cvpr}.

\subsection{Patch Similarity}
The most related work to confidence prediction is to compute the similarity between two patches \cite{BMVC2016_23,matchnet_cvpr_15,shaked2016stereo}.
The confidence score is learned by training the network with reflective loss in \cite{shaked2016stereo}.
The similarity is trained via a classification pipeline in \cite{BMVC2016_23}.
Patched representation as well as robust feature comparison is jointly learned in \cite{matchnet_cvpr_15}.

\begin{figure*}[pt]
	\centering	
	\includegraphics[width=0.87\textwidth]{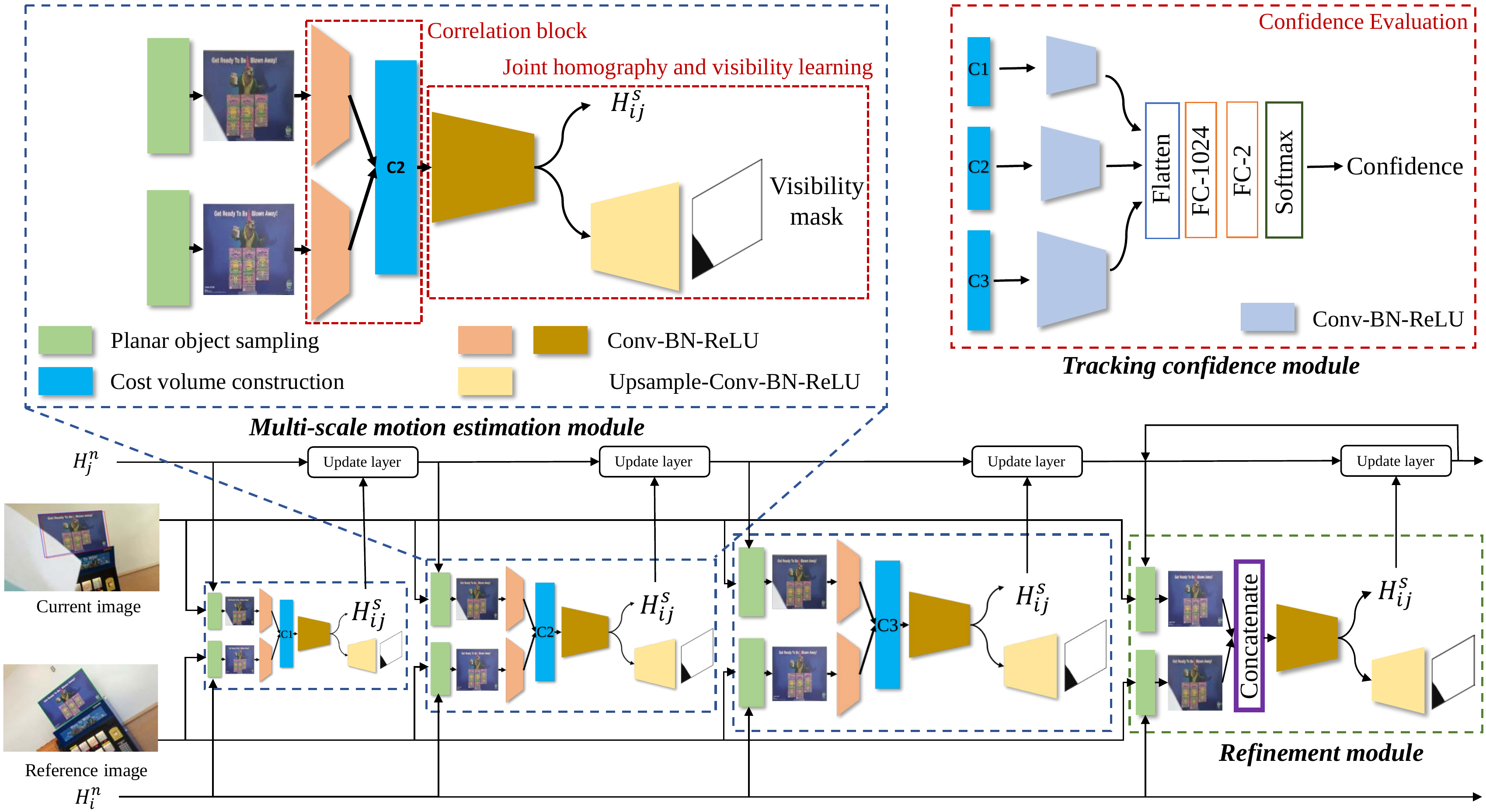}
	\caption[The framework of our approach.]{The framework of our model.
		It follows the LK scheme.
		There are three modules: the multi-scale motion estimation module, the refinement module, and the tracking confidence module.
		The base of this model is correlation blocks that extract features in the intensity domain for handling appearance variations and construct cost volumes in the pixel displacement domain for handling motion-related issues in a cascaded way (Sect.~\ref{subsec:decoupled_structure}).
		Pyramid blocks are build (Sect.~\ref{subsec:multi_scale_increments}), where homography and visibility are jointly learned (Sect.~\ref{subsec:joint_learning}).
		The refinement module for further improvements is optional (Sect.~\ref{subsec:refinement}).
		Tracking estimation confidence is also evaluated (Sect.~\ref{subsec:confidence}).
	}
	\label{fig:framework}

\end{figure*}

\section{Our Approach}
\label{sec:our_approach}
\subsection{The LK-based CNN Framework}
\label{subsec:pipeline}
Our model framework is shown in Fig.~\ref{fig:framework}.
We follows the LK scheme \cite{baker2004lucas} to compute homography, denoted as $\mathbf{H}_{ij} \in \mathbb{R}^{3\times3}$.
For each 3D object point $o_k$, its projection on image frame $i$ and $j$ is denoted as $\mathbf{p}_i^k$ and $\mathbf{p}_j^k$ respectively.
According to the derivation from \cite{richard2nd}, we have $\mathbf{p}_i^k = \mathbf{H}_{ij} \mathbf{p}_j^k$.
Supposing we have an initial homography $\mathbf{H}_{ij}$, the LK scheme consists of two iterated steps: \\
\noindent 1) solving for homography increment $\delta \mathbf{H}_{ij}$, \\
\noindent 2) updating homography $\mathbf{H}_{ij}  \leftarrow \mathbf{H}_{ij} * \delta \mathbf{H}_{ij}$. \\
For the first step, the classic LK method \cite{baker2004lucas} assume that intensities are consistent across images.
We improve this step with three aspects.
Firstly, as the intensity consistency assumption is prone to be broken in real-world cases with appearance variations and occlusions, we extend it with the `feature consistency' assumption and improve the effectiveness of feature representation (Sect.~\ref{subsec:decoupled_structure}).
Secondly, homography increments are computed with difference scales (Sect.~\ref{subsec:multi_scale_increments}).
Thirdly, based on the `feature consistency' assumption, we compute homography increments with joint homography and visibility learning (Sect.~\ref{subsec:joint_learning}).
The improved first LK step is implemented as the multi-scale motion estimation module in our model.
We also have an optional step without correlation block, i.e. the refinement module (Sect.~\ref{subsec:refinement}).
As computed homography increments are sensitive to homography initializations, we present a tracking confidence module to evaluate the estimation quality and re-initializes the homography computations (Sect.~\ref{subsec:confidence}).
We follow the same second step as the LK pipeline, where we update homography through update layers.
Lastly, we notice that the concerned planar object tracking problem is to solve for homography between object projections on two images while existing LK-based methods consider homography between two images.
We thus propose a sampling trick to turn the concerned problem into a classic LK-based homography problem that is more suitable for CNN models (Sect.~\ref{subsec:sampling}).

\subsection{Homography Surrogate \& Sampling}
\label{subsec:sampling}
\begin{figure}[!h]
	\centering
	\includegraphics[width=0.60\textwidth]{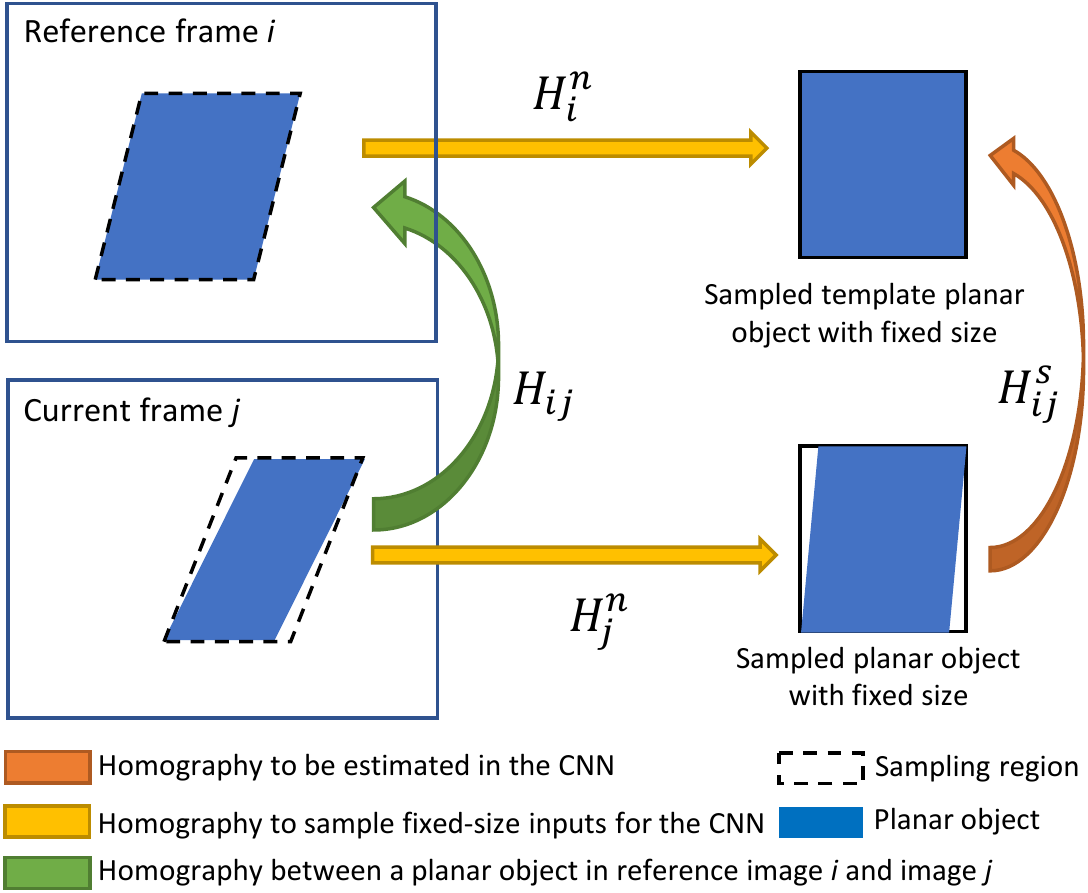}
	\caption[The planar object sampling layer.]{The planar object sampling.
		We sample the planar object in the reference frame and in the current frame to fixed-size images with $\mathbf{H}_{i}^n$ and $\mathbf{H}_{j}^n$, and use our mode to predict the increment $\mathbf{H}_{ij}^s$.
		$\mathbf{H}_{ij}^s$ will be the identity matrix if and only if the sampled planar objects on both sampled images are aligned perfectly.
		$\mathbf{H}_{j}^n$ and $\mathbf{H}_{ij}^s$ are used as surrogates for $\mathbf{H}_{ij}$ and $\delta \mathbf{H}_{ij}$ respectively.
	}
	\label{fig:sampling}

\end{figure}

The projection shape of a 3D plane on video images deforms as the camera moves relatively to the tracked object.
Processing the full-resolution video images with CNNs will waste a lot of memory as well as computations on useless image regions outside the projection shape.
What's worse, information on outside regions will distort the estimations and make CNN predictions more challenging.
To this end, we propose a planar object sampling layer for CNNs for handling planar objects in arbitrarily deformed shapes or sizes.
As shown in Fig.~\ref{fig:sampling}, the key idea is NOT to predict the original homography in the original image space.
Instead, we predict a surrogate homography in the normalized space.
We sample the planar object in the reference image into a $W \times H$ template: $\mathbf{p}_i^n = \mathbf{H}_{i}^n \mathbf{p}_i$, where $\mathbf{H}_{i}^n$ can be easily computed using SVD \cite{richard2nd} once the reference planar object with four-corner representation is given.
We denote the homography used to sample the planar object in the current image into a $W \times H$ template as $\mathbf{H}_{j}^n$, and the homography between two normalized images $i$ and $j$ is $\mathbf{H}_{ij}^s$. We have:
%{\small
	\begin{flalign}
		\mathbf{H}_{j}^n = (\mathbf{H}_{ij}^s)^{-1} \mathbf{H}_{i}^n \mathbf{H}_{ij}  = (\mathbf{H}_{ij}^s)^{-1} \mathbf{H}_{ij}^*
	\end{flalign}
	%}%
where $\mathbf{H}_{ij}^* = \mathbf{H}_{i}^n \mathbf{H}_{ij} $.
We define $\mathbf{H}_{j}^n$ as a surrogate for $\mathbf{H}_{ij}$, and $\mathbf{H}_{ij}^s$ as a surrogate for  $\delta \mathbf{H}_{ij}$.
$\mathbf{H}_{ij}^s$ will be an identity matrix if and only if $\mathbf{H}_{j}^n$ is equal to ground truth $\mathbf{H}_{ij}^*$.
If the final $\mathbf{H}_{j}^n$ is obtained, $\mathbf{H}_{ij}$ is computed as $\mathbf{H}_{ij} = (\mathbf{H}_{i}^n)^{-1}\mathbf{H}_{j}^n$.
By using surrogates, we maintain a fixed-size input to CNNs.

\subsection{Correlation Block}
\label{subsec:decoupled_structure}

Different from previous works \cite{aniel16,lin2017inverse} that regress homography on images, we decompose the homography regression into two cascaded steps:\\
\noindent 1) The first step is to extract features representing image local appearances.
These features are designed to be robust for image blur, illumination variations, occlusions, scale changes, perspective distortions, etc, through data argumentation covering various image conditions.
Since the template size is small, we use the U-Net structure \cite{u-net} for simplicity.
Other feature extraction structures, such as ResNet, EfficientNet and MultiResUNet, can also be used.\\
\noindent 2) The second step is to construct cost volumes with extracted features, whose elements are pixel correlations between sampled images.
These pixel correlations are designed to encode the relative geometry transformation between objects and cameras.
Each element in this cost volume is computed as the correlation \cite{pwc-net} between a pixel $x_i$ in reference feature map $\mathbf{f}_r$ and a pixel $x_j$ in the tracked feature map $\mathbf{f}_t$: $c(\mathbf{x}_i, \mathbf{x}_j ) = \mathbf{f}_r(\mathbf{x}_i) ^T \mathbf{f}_t(\mathbf{x}_j)$, where $T$ is the transpose operator.
Given a maximum displacement $d_m$, for each location $\mathbf{x}_i$ we compute correlations $c(\mathbf{x}_i, \mathbf{x}_j )$ for $\mathbf{x}_j$ s.t. $|\mathbf{x}_j - \mathbf{x}_i| <= d_m$.
Correlations at each location $\mathbf{x}_i$ are reorganized in the channel dimension.
Thus, the size of the 3D cost volume is $H \times W \times (2d_m+1)^2$.
$d_m$ is set to be 4 at each pyramid here by balancing the complexity and movement range.

\subsection{Pyramids}
\label{subsec:multi_scale_increments}

Inspired by the classic pyramid methods in image processing, we build correlation blocks in different scales.
We sample objects with different template resolutions (1/16x, 1/4x, 1x).
Homography increments are computed sequentially from the smallest resolution to the highest resolution.

\subsection{Joint Learning of Homography and Visibility}
\label{subsec:joint_learning}

Homography is obtained by information that is visible on both reference and tracked images.
Hence, we learn homography jointly with visibility, in order to extract a more reliable feature representation.
This leads to three loss functions during training: $L_d$, $L_m$, and $L_v$.
For benefit of CNNs, we adopt representation in \cite{aniel16}, where homography is represented by four corner displacements $\{d_1, d_2, d_3, d_4\}$.
$L_d$ is a homograph loss.
It is defined as the $l1$ norm between the ground truth 4-point displacement $d_k^*$ and the predicted 4-point displacement $d_k$ at each scale level:
{\small
\begin{flalign}
	{L_d} = \frac{1}{4}\sum\limits_{k = 1}^4 {{{\left\| {d_k^* - {d_k}} \right\|}_1}}
\end{flalign}
}%

$L_m$ is a visibility loss.
Pixel visibility prediction of the sampled tracked image is regarded as a 2-class classification problem.
We denote the ground truth label and the predicted label for a pixel's visibility as $m_k^*$ and $m_k$.
Cross-entropy is adopted for the visibility loss $L_m$ at each scale level:
{\small
	\begin{flalign}
		{L_m} =  - \frac{1}{N^k} \sum\limits_{k=1}^{N^k} (m_{k}^*\log (m_{k}^{}) + (1 - m_{k}^*)\log (1 - m_{k}^{}))
	\end{flalign}
}%
where $N^k$ is the total number of pixels at each scale level.
To further improve the feature representations used to construct cost volumes, we add a
visible alignment loss $L_v$ that minimizes the visible feature distance between extracted reference feature map $\mathbf{f}_r$ and tracked feature map $\mathbf{f}_t$.
It is defined as followed,
{\small
	\begin{flalign}
		{L_v} = \frac{1}{N^k} \sum\limits_{\mathbf{x}_k} m_{k}^* {\left\| { \mathbf{f}_t'(\mathbf{x}_k) - \mathbf{f}_t(\mathbf{x}_k) } \right\|}_1
	\end{flalign}
}%
where $\mathbf{x}_k$ is the pixel location on the sampled tracked image, $\mathbf{f}_t'= Warp(\mathbf{f}_r, \mathbf{H}_{tr} )$ is a wrapped feature map from $\mathbf{f}_r$ to $\mathbf{f}_t$ using the homography $\mathbf{H}_{tr}$.
The total loss is the combination of these three losses:
{\small
	\begin{flalign}
		L_{all}= {\lambda _d}{L_d} + {\lambda _m}{L_m} + {\lambda _v}{L_v}
	\end{flalign}
}%
where ${\lambda _d}$, ${\lambda _m}$ and ${\lambda _v}$ are balancing parameters.
In our experiments, they are all empirically set to be 1.0.

With the visibility loss, we explicitly connect homography with occlusion.
This is in contrast to competing methods \cite{clkn,nguyen2017unsupervised,lin2017inverse,zeng18rethinking,zhang2020content}  that handle occlusions implicitly with the learned feature capability.
Moreover, with the visible alignment loss, we ale able to connect homography, visibility and features in the correlation block.

Notice that, the supervised visibility mask varies in each scale level.
It is generated at each training iteration.

\subsection{Homography and Visibility Refinement}
\label{subsec:refinement}

This module is similar to that of Sect.~\ref{subsec:joint_learning} expect that the correlation block is removed and the visible alignment loss is ignored.
It is designed to capture tiny modifications to the homography and visibility.
The VGG structure \cite{vggnet} is used for simplicity.
Three iterations are usually conducted for convergence.
Note that, this module is optional.

\subsection{Estimation Confidence Evaluation}%\& Learning
\label{subsec:confidence}

This section discusses the homography initialization in the LK pipeline (Sect.~\ref{subsec:pipeline}).
The initial homography of the first scale level is equal to the homography obtained at the previous video frame $j-1$.
For the following scale levels, their initializations are equal to homography obtained at previous scale levels.
For the refinement module, its first homography initial value is equal to the homography output from the multi-scale motion estimation module.
In the following refinement step, its initial homography is equal to the homography in last iteration.%the last iteration.

With this homography initialization mechanism, we see the significance of the homography obtained at the previous video frame $j-1$, as it is the base of estimation in the current video frame $j$.
However, though we have tried our best to improve the homography estimation robustness and accuracy, our trained model inevitably fails under extreme conditions, such as large appearance variations, rapid camera-object relative motions, and severe occlusions.
That is, the homography obtained at the previous video frame $j-1$ may be unreliable.
To check this, we add a tracking confidence module to evaluate the estimation confidence.
This confidence is regarded as a regression whose output ranges between 0 and 1.
0 indicates the estimation is unreliable while 1 indicates it is reliable.
In contrast to previous works \cite{BMVC2016_23,matchnet_cvpr_15,shaked2016stereo} that regress confidences from images, we regress them from cost volumes of correlation blocks.
These multi-scale cost volumes, representing distributions of pixel correlations, encode the `uncertainty' of the estimation.
For an object pixel in the reference image, its corresponding pixel on the tracked image is ambiguous if the pixel correlation distribution is flat, or obvious if the pixel correlation distribution is concentrated on one specific location.
We train this tracking confidence module after the multi-scale motion estimation module and the optional refinement module is trained using an independent dataset.

We consider the estimation as unreliable if the homography loss $L_d$ between the ground truth and predicted homography is larger than 5 while reliable otherwise.
We denote the ground truth label and the predicted label as $p^*$ and $p$.
Cross-entropy loss is used for confidence loss:
{\small
	\begin{equation}
		{L_c} =  - ({p^*}\log (p) + (1 - {p^*})\log (1 - p))
	\end{equation}
}%
In implementations, each cost volume of each pyramid layer is convoluted to a $\frac{H}{8} \times \frac{W}{8} \times 15$ feature map by several convolutional layers respectively.
These feature maps are then followed by two fully connected (FC) layers, whose drop-out ratio is set to 0.5, with 1024 and 2 channels.
The final layer is a soft-max layer that output the confidence.
$3 \times 3$ kernels are used in convolutional layers.

After the tracking confidence module is trained, we monitor the tracking confidence on the fly.
If the homography obtained at the previous video frame $j-1$ is classified as unreliable, we use the homography estimated in more previous times (e.g. 2 to 60 frames before) for homography initialization and re-run our model pipeline.
This process is repeated until this tracking is reliable.

\section{Experiments}
\label{sec:exp}
\begin{figure}[t]%[!h]
	\centering
	\includegraphics[width=0.5\columnwidth]{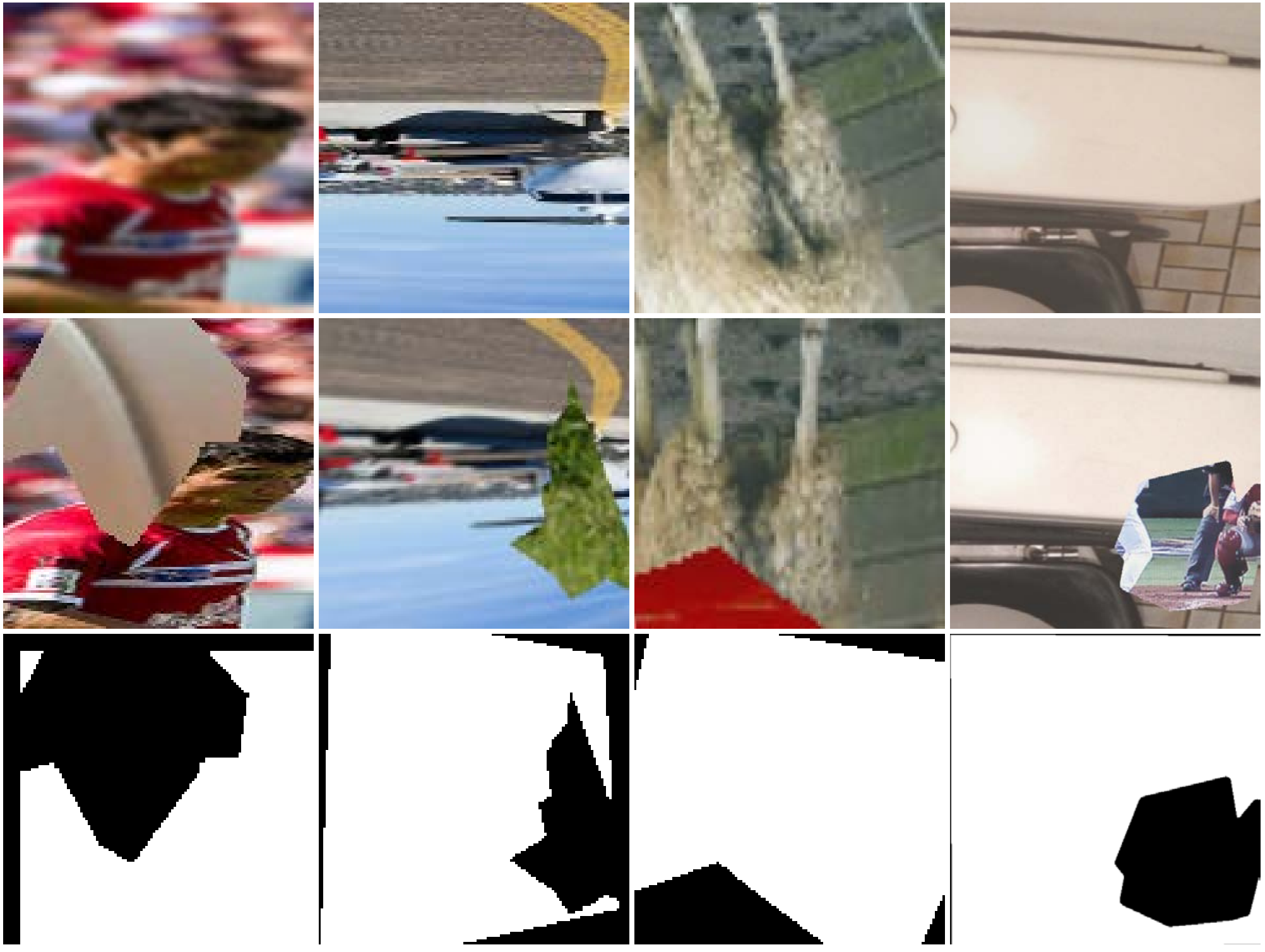}
	\caption{Samples of the generated dataset.
		First two rows are image pairs with variations of brightness, contrast, saturation, image blur, and occlusions.
		The last row shows ground truth visibility masks. %GT
	}
	\label{fig:dataset_sample}

\end{figure}

Similar to \cite{clkn,aniel16}, we use the MS-COCO dataset \cite{COCO14} to generate the training data.
All images are resized to $240 \times 240$.
We randomly select an image, assign a $120 \times 120$ window to its center. %$128 \times 128$
We then randomly perturb the four corners of this window to generate a random homography.
The corner displacement is uniformly distributed between [-32, 32] in both horizontal and vertical directions.
Pixels within the perturbed window are wrapped to a sample image whose size is $W \times H$.
To increase the robustness of our network, we augment our samples with more conditions that we meet in real-world applications.
We add variances of brightness, contrast, saturation and image blur to the sample images \cite{data_augmentation}.
Moreover, we simulate real-world object occlusions by randomly placing arbitrary polygons, whose textures are cropped natural images from \cite{COCO14}, into our training samples \cite{data_augmentation}.
280000 image pairs with ground truth homography are generated in total (Fig.~\ref{fig:dataset_sample}).
Among them, 200000 samples are used for training the motion estimation network and refinement network, 40000 samples are used for validation, and the rest 40000 samples are tested for ablation study (Sect.~\ref{subsec:ablation}).
GT visibility masks are generated at each training iteration.

\label{subsec:training}

\subsection{Training \& Quantitative Evaluation}
In all experiments, we set $W = H = 120$.
Adam~\cite{adam} optimization with ${\beta _1=0.9}$, ${\beta _2=0.999}$ is used, and the batch size is set to 32.
Batch normalization~\cite{bn} is adopted for accelerating convergence.
The learning rate is initialized to be $10^{-4}$.
It is then decreased by a factor of 10 every 5 epochs.
After the model is trained, its processing rate is about 10hz on a commodity GPU card GeForce GTX 1080.	

In this paper, two quantitative metrics, alignment error (AE)~\cite{TMT_benchmark} and homography discrepancy (HD)~\cite{planar_tracking_benchmark}, are used to evaluate the quality of predicted homography accuracy.

\subsection{Ablation Study}
\label{subsec:ablation}	
In this section, we perform ablation studies to analyze the contribution of each component in our proposed model.
All methods are trained on the training dataset as well as tested on the dataset from Sect.~\ref{subsec:training} introduction.

\subsubsection{Homography Precision}
	
We firstly analyze component contributions to the homography precision.
We train our model with increasing components proposed in this paper: the correlation block in Sect.~\ref{subsec:decoupled_structure} (D), pyramids in Sect.~\ref{subsec:multi_scale_increments} (P), joint learning of homography and visibility in Sect.~\ref{subsec:joint_learning} (M), the refinement module in Sect.~\ref{subsec:refinement} (R): Ours-D, Ours-DP, Ours-DPR, Ours-DPM, Ours-DPMR.
If our model is trained without any proposed components (Ours w/o DPMR), it is equivalent to DeepHomography \cite{aniel16}.
Tab.~\ref{tab:decoupled} shows the results:

\begin{table}[!h]
	\centering

	\caption{Ablation study and comparison on our test set.}	
	{	
		\footnotesize
        \setlength{\tabcolsep}{18.5pt}
        \renewcommand{\arraystretch}{0.9}
		\begin{tabular}{l|cc}
			\toprule
			Method & AE~\cite{TMT_benchmark} & HD~\cite{planar_tracking_benchmark} \\
			\midrule
			%IC-STN \cite{lin2017inverse} & 2.855 & 5.850 \\
			Ours w/o DPMR & 6.678 & 14.983  \\
			Ours-P & 5.280 & 10.627  \\
			Ours-PR  & 2.970 & 5.104  \\			
			Ours-PM  & 4.051 & 7.984  \\			
			Ours-PMR  & 2.426 & 4.262  \\			
			Ours-D & 4.173 & 9.147  \\
			Ours-DP & 1.145 & 2.216  \\
			Ours-DPR &  \textbf{\textcolor{blue}{0.876}} & 1.739  \\		
			Ours-DPM & 1.097 & 2.107  \\				
			Ours-DPMR &  \textbf{\textcolor{blue}{0.876}} &  \textbf{\textcolor{blue}{1.695}}  \\			
			\bottomrule
		\end{tabular}
	}
	\label{tab:decoupled}

\end{table}

\noindent -- From line 2 and line 7, we see that the model with correlation blocks (Ours-D) performs significantly better than that without them (Ours w/o DPMR). \\
\noindent -- Pyramids (P) do help both approaches (Ours-D and Ours w/o DPMR).
This improvement is more significant for the model Ours-D as the cost volume is constructed on limited displacements. \\
\noindent -- The refinement module is able to capture tiny displacement between images.
It further increases the accuracy for all models (Ours-DP vs Ours-DPR, Ours-DPM vs Ours-DPMR, Ours-P vs Ours-PR, Ours-PM vs Ours-PMR). \\
\noindent -- By jointly training homography and visibility, our model generalizes better on each original task (Ours-DP vs Ours-DPM, Ours-P vs Ours-PM, and Ours-PR vs Ours-PMR).

\subsubsection{Visibility Accuracy}

\begin{figure}[t]
	\centering	
	\includegraphics[width=0.55\textwidth]{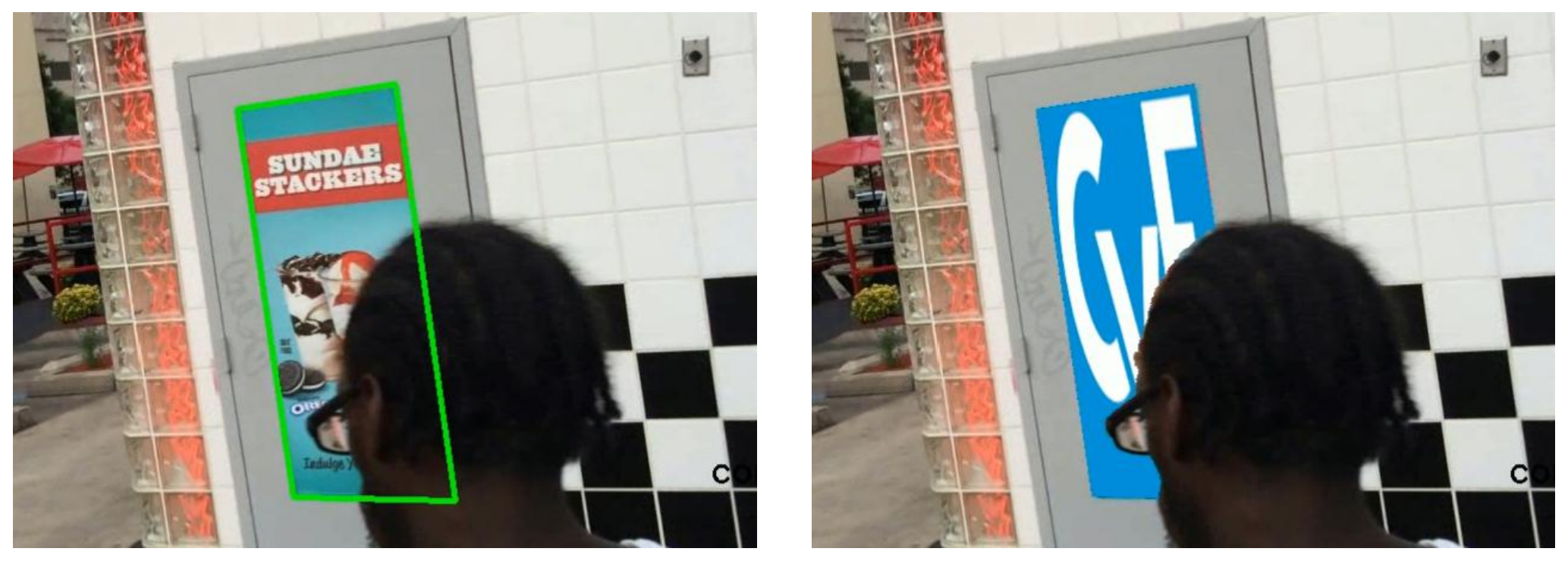}
	\caption{A challenging case with large and irregular occlusions.
	}
	\label{fig:occlusion}

\end{figure}

\begin{table}[t]

	\caption{Visibility loss of models w/ or w/o the correlation block (D), w/ or w/o joint homography and visibility learning (M vs V).}
	\begin{adjustbox}{width=0.75\columnwidth,center}
		{
			\scriptsize
            %\footnotesize
			\begin{tabular}{c|cccc}
				\toprule
				Method & Ours-PVR &Ours-PMR & Ours-DPVR & Ours-DPMR\\
				\midrule
				Visibility loss & 0.347& 0.346 & 0.335 & \textbf{\textcolor{blue}{0.328}} \\
				\bottomrule
			\end{tabular}
		}	
	\end{adjustbox}
	\label{tab:mask_loss}

\end{table}

Apart from the improvement to homography precision, we wonder whether learning of homography and visibility jointly (M) leads to higher visibility accuracy than learning these two tasks independently (V).
We also test if the correlation block helps visibility accuracy.
We train four models on the generated training dataset: Ours-PMR, Ours-DPMR, Ours-PVR and Ours-DPVR.
We then compute the visibility loss (Sect.~\ref{subsec:joint_learning}) on the test set.
Results are shown in Tab.~\ref{tab:mask_loss}.
We find that the correlation block and joint learning not only help the homography predictions but also improve the visibility estimations.
We see strong connections between homography and visibility.
Visibility, a by-product of our work, can be used for in-video advertising.
We show one synthesized frame (Fig.~\ref{fig:occlusion}) using our obtained visibility during experiments on the POT dataset \cite{planar_tracking_benchmark}.
We meet large and irregular occlusions that are challenging to our model.
Fortunately, our model is able to overcome this difficulty.

\subsubsection{Confidence Effectiveness}
\label{exp:confidence}
One way to evaluate the confidence effectiveness is to compute the classification statistics using the predicted confidence (0.5 is used as the threshold).
We follow data generations in Sect.~\ref{subsec:training} introduction to generate an additional large dataset covering challenging conditions.
This dataset, on which tracking is much harder than that of in Sect.~\ref{subsec:training} introduction, contains 50000 samples.
The percents of training, validation and testing are 80\%, 10\% and 10\% respectively.
Our trained models (Ours-DPR and Ours-DPMR) are then run on this dataset.
If the computed $L_d$ is smaller than 5, the tracking result is labeled to be reliable.
Otherwise, it is labeled to be unreliable.
Obtained labels are adopted for training the confidence network and testing the confidence performance.
PatchCon \cite{BMVC2016_23} that directly regresses this confidence from wrapped images is the baseline/competing method. %input
Both OursCon and PatchCon are trained to evaluate pre-trained Ours-DPMR and Ours-DPR.

True-positive rate (TPR), false-positive rate (FPR), false-negative rate (FNR) and true-negative rate (TNR) are shown in Tab.~\ref{tab:confidence}.
Comparing OursCon and PatchCon \cite{BMVC2016_23} that both evaluate Ours-DPMR, we see that tracking confidence predicted from correlation blocks is more accurate.
Moreover, from OursCon +Ours-DPR and OursCon+Ours-DPMR, we see that joint learning of visibility mask and homography does improve the effectiveness of our correlation block and model generalization, leading to performance gains of confidence prediction.

\begin{table}[t]%!h

	\caption{Classification statistics using the estimated confidence.}
	\centering
    \setlength{\tabcolsep}{9pt}
    \renewcommand{\arraystretch}{0.7}
	{
        \tiny
        \scriptsize
		%\footnotesize
		\begin{tabular}{c|cccc}
			\toprule
			Method + Pre-trained Base & TPR & FPR & FNR & TNR \\
			\midrule
			PatchCon \cite{BMVC2016_23}+Ours-DPMR  & 93.1\% &14.5\%&  6.9\% & 85.5\% \\			
			OursCon+Ours-DPMR &  \textbf{\textcolor{red}{96.6\%}} &  \textbf{\textcolor{blue}{8.7\%}}&   \textbf{\textcolor{blue}{3.4\%}} &  \textbf{\textcolor{red}{91.3\%}} \\
			OursCon+Ours-DPR & 96.5\% &  10.2\%&  3.5\% & 89.8\% \\
			\bottomrule
		\end{tabular}
	}
	\label{tab:confidence}

\end{table}

\begin{figure*}[t]%[htbp]%

	\centering
	\includegraphics[width=0.96\textwidth]{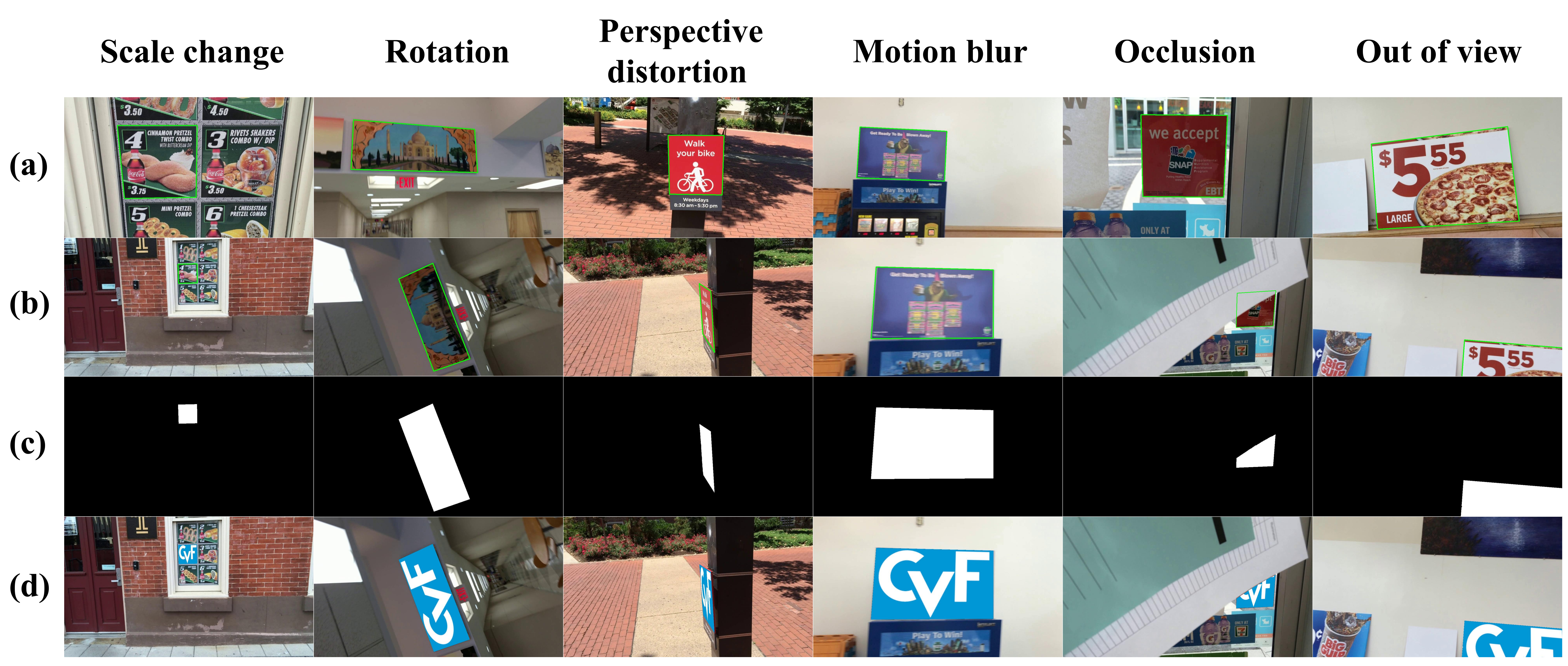}
	\caption{Results obtained by our model in different conditions. (a) A planar object in the reference frame. (b) The tracked planar object in the current frame.
		(c) Predicted visibility mask corresponding to (b). (d) The synthetic frame after placing the CVF logo on (b).
		More results can be found in the supplementary material.
	}
	\label{fig:results}

\end{figure*}

\subsection{Comparisons on Other Datasets}

Two public datasets, POT~\cite{planar_tracking_benchmark} and TMT~\cite{TMT_benchmark}, are used to evaluate the homography accuracy.
State-of-the-art methods, including SIFT~\cite{sift}, SURF~\cite{surf}, L1~\cite{L1APG,L1_track}, IVT~\cite{IVT}, ESM~\cite{esm04}, Gracker~\cite{gracker}, DeepHomography~\cite{aniel16}, IC-STN~\cite{lin2017inverse}, Ctx-Unsupervise~\cite{zhang2020content}, PFN~\cite{zeng18rethinking}, MHN~\cite{Le_CVPR_2020} and DLKFM~\cite{zhao2021deep} are compared.
Our models are all with our tracking confidence module (OursCon), except the one named Ours-DPMR w/o OursCon.
The competing confidence prediction method, PatchCon \cite{BMVC2016_23}, is also included for comparison (Ours-DPMR-PatchCon).
The model with all our modules achieves the best performance.

\begin{figure*}[htpb]
	\centering
	\includegraphics[width=0.95\textwidth]{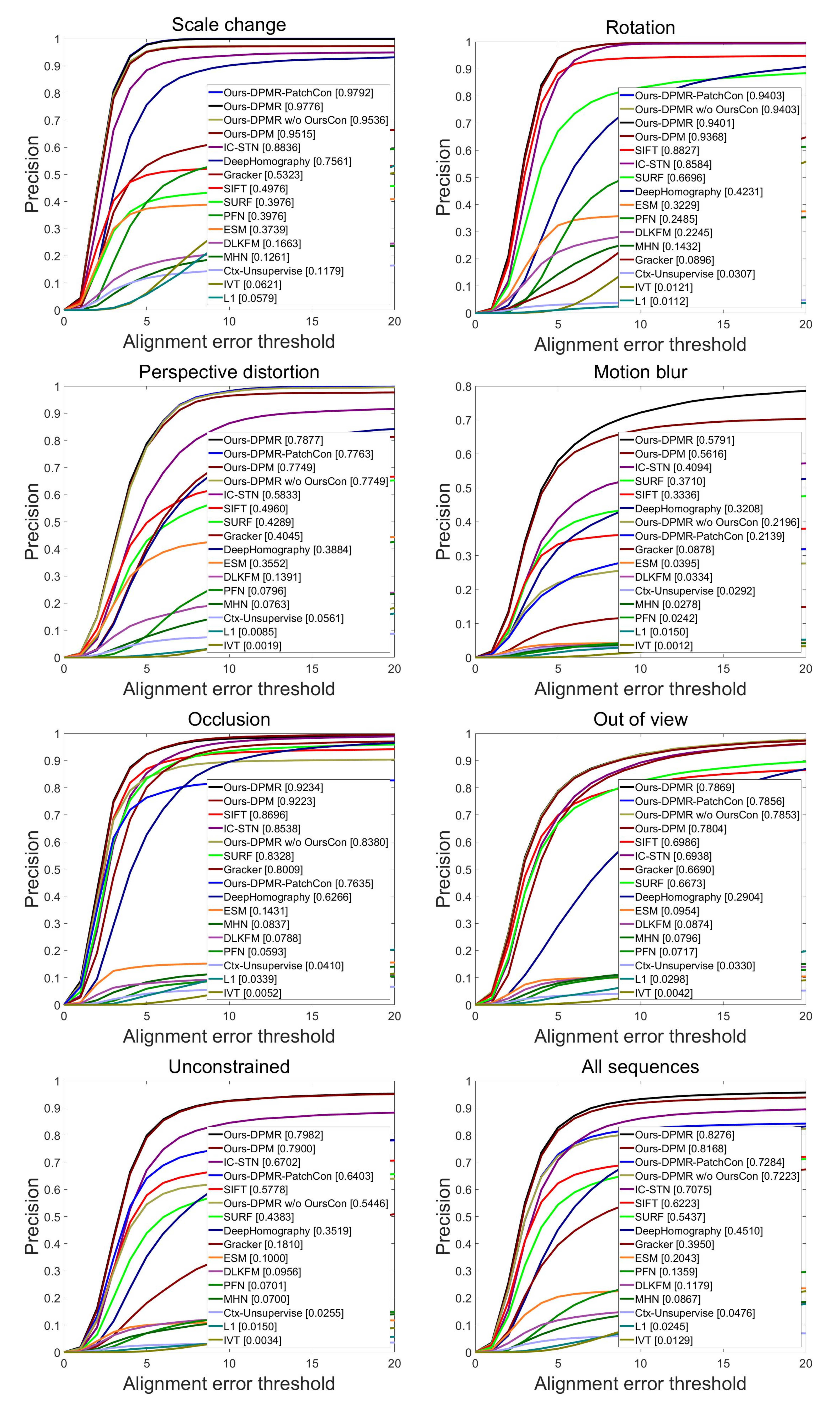} %height=3.25cm% Reduce the figure size so that it is slightly narrower than the column. Don't use precise values for figure width.This setup will avoid overfull boxes.
	\caption{The comparison of different approaches shown in precision plots on the POT dataset \cite{planar_tracking_benchmark}.
		Curves with larger areas are better.
		The AE at threshold $=5$~\cite{TMT_benchmark} is illustrated within brackets.
		Zoom-in is recommended.
		Video comparisons are in the supplementary material.
	}
	\label{fig:precision_plot}

\end{figure*}

\noindent \textbf{POT } is a planar object tracking benchmark containing 210 videos of 30 planar objects in natural environments.
It contains scenes with various challenging conditions, including scale change, rotation, perspective distortion, motion blur, occlusion, out-of-view, and a combination of these factors.
For better presentation, comparisons are shown with precision plots and success plots.
Precision plot counts the percentage of frames whose AE is within the threshold $t_p$.
Success plot counts the percentage of frames whose HD is within a threshold $t_s$.
Results are shown in Fig.~\ref{fig:precision_plot} and the supplementary material.
Our proposed method shows superior performance in all scenes.
Especially for scenes with motion blur, perspective distortion, scale change or combinations of these factors, our approach works much better because it is hard for non-learning algorithms to model the underlying variation or tuning related parameters manually.

\noindent \textbf{TMT } consists of sequences for manipulation tasks.
There are 100 annotated and tagged sequences in total.
Similar to POT, sequences in this dataset also have a large condition variation.
%We select sequences whose objects to be tracked are planar or approximately planar.
We use the same evaluation metric as in \cite{TMT_benchmark}.
That is, the success rate that counts the percentage of frames whose AE $< 5$.
Comparison results are summarized in Table.~\ref{tab:TMT}.
Overall, our model achieves a better or similar performance in all sequences compared to other methods.

\begin{table*}[t]%[htbp]
    \caption{Success rate of different approaches on the TMT dataset with AE $< 5$~\cite{TMT_benchmark}.
	Larger is better.
	Best and second best are colored. \ \
	(*) Models of Ours-DPMR, Ours-DPMR-PatchCon and Ours-DPMR w/o OursCon perform the same.
	We omit rested notations for short.
	}
	\centering
    \setlength{\tabcolsep}{2.0pt}
    \renewcommand{\arraystretch}{1.0}
	{
        \tiny
        \scriptsize
		%\footnotesize
		\begin{tabular}{c|cccccccccccc}%{c|c|c|c|c|c|c|c|c|c|c|c|c} %{c|cccccccccccc}
			\toprule
			\scriptsize Method & \tiny Cereal & \tiny Book1 & \tiny Book2 & \tiny Book3 & \tiny Juice & \tiny Mug1 & \tiny Mug2 & \tiny Mug3 & \tiny Bus  & \tiny Highlight & \tiny Letter & \tiny Newspaper \\ %Highlighting
            %\scriptsize Method &  Cereal &  Book1 &  Book2 &  Book3 &  Juice &  Mug1 &  Mug2 &  Mug3 &  Bus  &  Highlighting &  Letter &  Newspaper \\
			\midrule
			SIFT~\cite{sift} & 0.92 & \textcolor{green}{0.74} & \textcolor{red}{1.00} & 0.84 & 0.89 & 0.91 & 0.43 & 0.55 & 0.19 & 0.97 & 0.18 & 0.16 \\
			SURF~\cite{surf} & 0.91 & 0.64 & \textcolor{red}{1.00} & 0.74 & 0.50 & 0.07 & 0.14 & 0.06 & 0.19 & 0.94 & 0.08 & 0.01 \\								
			L1~\cite{L1_track} & 0.24 & 0.10 & 0.79 & 0.42 & 0.16 & 0.10 & 0.30 & 0.54 & 0.57 & 0.67 & 0.19 & 0.61 \\
			IVT~\cite{IVT} & \textcolor{green}{0.99} & 0.48 & 0.30 & 0.72 & 0.98 & 0.91 & 0.72 & 0.68 & 0.94  & 0.95 & 0.25 & 0.92 \\
			ESM~\cite{esm04} & \textcolor{red}{1.00} & \textcolor{red}{1.00} & \textcolor{red}{1.00} & 0.34 & \textcolor{red}{1.00} & \textcolor{red}{1.00} & \textcolor{red}{0.89} & \textcolor{red}{1.00}& \textcolor{red}{1.00} & 0.76 & \textcolor{red}{1.00} & \textcolor{red}{1.00} \\
			%% NNBMIC~\cite{NNBMIC} & \textcolor{red}{1.00} & \textcolor{red}{1.00} & \textcolor{red}{1.00} & 0.32 & 0.41  & \textcolor{red}{1.00} & \textcolor{red}{0.89} & 0.59 & 0.99 & 0.70 & \textcolor{red}{1.00} & 0.51 \\
			%% BMIC~\cite{BMIC} & \textcolor{red}{1.00} & \textcolor{red}{1.00} & \textcolor{red}{1.00} & 0.32 & \textcolor{red}{1.00} & \textcolor{red}{1.00} & \textcolor{red}{0.89} & 0.65 & 0.96 & 0.33 & \textcolor{red}{1.00} & 0.43 \\
			Gracker \cite{gracker} & 0.91 & \textcolor{red}{1.00} & \textcolor{red}{1.00} & \textcolor{red}{0.88} & \textcolor{red}{1.00} & \textcolor{red}{1.00} & 0.83 & 0.75 & 0.97 & \textcolor{red}{1.00} & \textcolor{green}{0.78}& \textcolor{red}{1.00} \\				
			%DeepHomography \cite{aniel16} & 0.92 & \textcolor{red}{1.00} & \textcolor{red}{1.00} & 0.82 & \textcolor{green}{0.99} & 0.93 & 0.65 & 0.80 & 0.50 & \textcolor{green}{0.99} & \textcolor{red}{1.00} & 0.95 \\
            DeepHomography & 0.92 & \textcolor{red}{1.00} & \textcolor{red}{1.00} & 0.82 & \textcolor{green}{0.99} & 0.93 & 0.65 & 0.80 & 0.50 & \textcolor{green}{0.99} & \textcolor{red}{1.00} & 0.95 \\
			IC-STN \cite{lin2017inverse} & 0.92 & \textcolor{red}{1.00} & \textcolor{red}{1.00} & 0.82 & \textcolor{red}{1.00} & \textcolor{red}{1.00} & 0.77 & 0.79 & \textcolor{green}{0.99} & 0.98 & \textcolor{red}{1.00} & 0.95 \\				
			PFN \cite{zeng18rethinking} & 0.74 & 0.28 & 0.92 & 0.38 & 0.39 & 0.89 & 0.40 & 0.88 & 0.24 & 0.78 & 0.29 & 0.53 \\				
			%Ctx-Unsupervise \cite{zhang2020content} & 0.54 & 0.38 & \textcolor{red}{1.00} & 0.38 & 0.29 & 0.28 & 0.23 & 0.39 & 0.16 & \textcolor{red}{1.00} &0.17 & 0.14 \\	
            Ctx-Unsupervise  & 0.54 & 0.38 & \textcolor{red}{1.00} & 0.38 & 0.29 & 0.28 & 0.23 & 0.39 & 0.16 & \textcolor{red}{1.00} &0.17 & 0.14 \\	
			MHN \cite{Le_CVPR_2020} & 0.62 & 0.18 & 0.92 & 0.40 & 0.63 &  \textcolor{green}{0.99} & 0.50 & 0.41 & 0.50 & 0.76 &0.22 & 0.14 \\
			DLKFM \cite{zhao2021deep} & 0.58 & 0.18 & 0.92 & 0.41 & 0.63 &  \textcolor{green}{0.99} & 0.50 & 0.41 & 0.50 & 0.76 & 0.21 & 0.14 \\			
			\midrule
			Ours-D & 0.85 & 0.65 & \textcolor{green}{0.84} & 0.67 & 0.37 & \textcolor{red}{1.00} & 0.78 & 0.72 & 0.71 & 0.92 & 0.50 & 0.32 \\
			Ours-DP & 0.93 & \textcolor{red}{1.00} & \textcolor{red}{1.00} & \textcolor{green}{0.86} & \textcolor{red}{1.00} & \textcolor{red}{1.00} & \textcolor{green}{0.84} & 0.81 & 0.97 & \textcolor{red}{1.00} & \textcolor{red}{1.00} & 0.93 \\
			Ours-DPR & 0.93 & \textcolor{red}{1.00} & \textcolor{red}{1.00} & \textcolor{red}{0.88} & \textcolor{red}{1.00} & \textcolor{red}{1.00} &  0.83 & 0.80 & 0.95 & \textcolor{red}{1.00} & \textcolor{red}{1.00} &  \textcolor{green}{0.98} \\
			Ours-DPM & 0.93 & \textcolor{red}{1.00} & \textcolor{red}{1.00} & \textcolor{red}{0.88} & \textcolor{red}{1.00} & \textcolor{red}{1.00} & 0.72 & 0.83 & \textcolor{green}{0.99} & \textcolor{red}{1.00} & \textcolor{red}{1.00} & 0.93 \\
			Ours-DPMR (*) & 0.93 & \textcolor{red}{1.00} & \textcolor{red}{1.00} & \textcolor{red}{0.88} & \textcolor{red}{1.00} & \textcolor{red}{1.00} & \textcolor{red}{0.89} & \textcolor{green}{0.85} & \textcolor{green}{0.99} & \textcolor{red}{1.00} & \textcolor{red}{1.00} & \textcolor{red}{1.00} \\
			\bottomrule
		\end{tabular}
	}
	\label{tab:TMT}
\end{table*}

We visualize some qualitative results obtained by our model during experiments and place a product (i.e. the CVF logo) on the tracked planar object in Fig.~\ref{fig:spot} and Fig.~\ref{fig:results}.
More results can be found in the supplementary material.	

\section{Discussions \& Limitations \& Conclusions}
\label{sec:conclusion}
The main limitation of our work is that the predicted visibility mask is not perfect.
With the own constraints of LK-based methods, our approach is sometimes disturbed by the factor of similar occluded objects.
%It fails if the occlusion looks very similar to the tracked object or if the jointly predicted homography is not accurate.
%To conclude this work, 
In conclusion, we proposed a novel model for planar object tracking.
Homography, visibility and confidence are jointly learned based on a correlation block.
We achieved a superior planar tracking performance compared to state-of-the-art methods on the public dataset, provided visibility masks that other works had not discussed, calculated more reliable confidence than competing approaches. % on our generated dataset.
To better take multi-frame constraints and similar occlusions into consideration is our future work.

\clearpage
\bibliographystyle{splncs04}
%\bibliography{egbib}
\bibliography{ref}
\end{document}